%% file: root.tex
\documentclass[letterpaper, 10 pt, conference]{ieeeconf}  

\IEEEoverridecommandlockouts                              

\usepackage{graphics}
\usepackage{epsfig} 
\usepackage{mathptmx} 
\usepackage{times} 
\usepackage{amsmath} 
\usepackage{amssymb}  
\usepackage{multirow}
\usepackage{url}
\usepackage{caption}
\usepackage{subcaption}
\usepackage[table]{xcolor}
\usepackage{import}
\usepackage{hyperref}
\usepackage{cleveref}
\usepackage{comment}
\usepackage{booktabs}
\usepackage{tabularx}
\usepackage{microtype}
\usepackage{gensymb} 
\usepackage{cite}
\usepackage{stfloats}
\usepackage{bm}
\usepackage{siunitx}
\usepackage{acro}
\DeclareAcronym{com}{ 
    short = {CoM}, 
    long  = {Center of Mass},
    first-style = short-long,
}
\DeclareAcronym{cog}{ 
    short = {CoG}, 
    long  = {Center of Geometry},
    first-style = short-long,
}

\DeclareAcronym{wrt}{ 
    short = {w.r.t.}, 
    long  = {with respect to},
    first-style = short-long,
}
\DeclareAcronym{ee}{ 
    short = {EE}, 
    long  = {End-Effector},
    first-style = short-long,
}
\DeclareAcronym{dof}{ 
    short = {DoF}, 
    long  = {Degree of Freedom},
    first-style = short-long,
}
\usepackage{csquotes}
\newcommand{\update}[1]{\textcolor{black}{#1}}

\title{\LARGE \bf \update{Dynamic Center-of-Mass Displacement in Aerial Manipulation: An Innovative Platform Design}}

\author{ Tong Hui${^1}^*$, Stefan Rucareanu$^1$, Esteban Zamora$^1$, Simone D'Angelo$^2$,\\
Haotian Liu$^1$, Matteo Fumagalli$^1$
\thanks{This work has been supported by the European Unions Horizon 2020 Research and Innovation Programme AERO-TRAIN under Grant Agreement No. 953454. $^*$ Corresponding author email: {\tt\small tonhu@dtu.dk}}
\thanks{${}^1$ Technical University of Denmark, Denmark}
\thanks{${}^2$ PRISMA Lab, Department of Electrical Engineering and Information Technology, University of Naples Federico \uppercase\expandafter{\romannumeral2}}
}

\begin{document}

\maketitle
\thispagestyle{empty}
\pagestyle{empty}

\begin{abstract}
\update{Aerial manipulators are increasingly used in contact-based industrial applications, where tasks like drilling and pushing require platforms to exert significant forces in multiple directions. To enhance force generation capabilities, various approaches, such as thrust vectoring and perching, have been explored. In this article, we introduce a novel approach by investigating the impact of varied \ac{com} locations on an aerial manipulation system's force exertion. Our proposed platform features a design with a dynamically displacing \ac{com}, enabling a smooth transition between free flight and high-force interactions supported by tilting back rotors. We provide detailed modeling and control strategies for this design and validate its feasibility through a series of physical experiments. In a pushing task, the proposed system, weighing $3.12\si{\kilogram}$, was able to stably exert over $28\si{\newton}$ of force on a work surface—nearly equivalent to its gravitational force—achieved solely through the tilting of its back rotors. Additionally, we introduce a new factor to evaluate the force generation capabilities of aerial platforms, allowing for a quantitative comparison with state-of-the-art systems, which demonstrates the advantages of our proposed approach.
}
\end{abstract}

\import{}{texts/01_introduction}
\import{}{texts/02_system_design}

\import{}{texts/03_modeling}
\import{}{texts/04_control}
\import{}{texts/06_experiment}
\import{}{texts/07_conclusion}
\import{}{texts/biblio}
\addtolength{\textheight}{-12cm}   






% ...

\end{document}

%% file: texts/01_introduction.tex
\section{Introduction} \label{sec:intro}
\update{Critical working conditions in industrial applications, such as oxygen-deficient environments, operations at heights, and offshore tasks, often pose challenges related to workspace safety, labor shortages, and high costs \cite{construct}. Recently, the utilization of aerial manipulators for such operations has become increasingly established \cite{review,review_0}. These platforms have demonstrated effectiveness in contact-based inspections through pushing  \cite{bart,truj,bodie2019,watson2022,tong_voliro,icra_tong_dtu,icra_tong_asl} and in maintenance operations like drilling and grinding \cite{sun,drill,ding,grinding}. Such tasks demand the aerial platform to exert significant forces in various directions.}
\begin{figure}[!t]
   \centering
\includegraphics[width=\columnwidth]{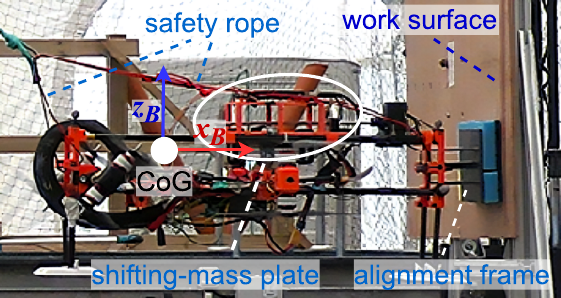}
    \caption{A still frame from the physical experiment with the built prototype. The body frame $\mathcal{F}^B=\{O_B;\bm{x}_B,\bm{y}_B,\bm{z}_B\}$ is attached to the aerial vehicle \ac{cog}. The aerial vehicle pushes on a vertical surface using an alignment frame along the body axis $\bm{x}_B$ with tilted back rotors. The system's \ac{com} is displaced by moving the shifting-mass plate toward the work surface while preserving the orientation and contact using the alignment frame.}
    \label{fig:physical_model}
    \vspace{-0.5cm}
\end{figure}

\update{The development of aerial manipulators, especially multirotors, capable of physically interacting with the environment has experienced significant growth over the past decade \cite{review_0,review}. Traditional uni-directional thrust \cite{review} multirotors (e.g., quadrotors, hexacopters, octocopters) are energy-efficient but are limited by the inherent coupling between linear and angular dynamics, which restricts their force generation capability in both magnitude and direction during interactions \cite{tong_aim,icra_tong_dtu}.}

\update{To address this, higher-\ac{dof} manipulators have been mounted on uni-directional thrust multirotors, as seen in \cite{bart,kim,tong_aim,icra_tong_dtu}. However, these platforms still face limitations in force magnitude due to the coupling between gravity compensation and horizontal force generation during physical interactions \cite{icra_tong_dtu}. An alternative is thrust vectoring, achieved either by fixed tilted rotors \cite{truj,antonio,odar}, or tiltable rotors \cite{mina,watson2022,ding,hwang,bodie2019}. In \cite{mina,truj,antonio,odar, watson2022,bodie2019,hwang}, authors present aerial vehicles achieving 6-\ac{dof} wrench generation in the space being fully-actuated.
These platforms often suffer from counteracting unwanted thrust components, high energy consumption, and control complexity due to extra actuation compared to uni-directional thrust multirotors. Moreover, full actuation is not always necessary for tasks like pushing and drilling on non-horizontal surfaces.}

\begin{figure}[!t]
   \centering
\begin{subfigure}{0.44\columnwidth}\centering
    \includegraphics[trim={0.8cm 0.4cm 0.8cm  0.0cm},clip,width=\columnwidth]{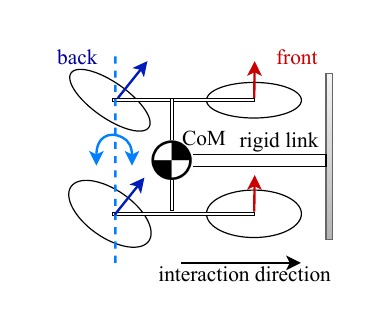}
    \caption{}
    \label{fig:design1}
\end{subfigure}
     \begin{subfigure}{0.54\columnwidth}\centering
    \includegraphics[trim={0.8cm 0.5cm 0.7cm  0.5cm},clip,width=\columnwidth]{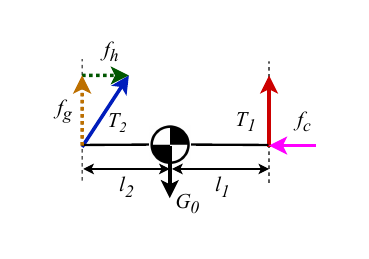}
    \caption{}
    \label{fig:design2}
\end{subfigure}
\caption{\update{(a) A simplified illustration of an H-shaped aerial vehicle with tiltable back rotors and fixed front rotors pushing on a vertical surface using the attached rigid link. (b) Free body diagram of the system in (a).}}\label{fig:design_motivation}
\end{figure}

\update{ Some solutions, such as the H-shaped aerial vehicles \cite{ding,hwang}, use 5-\ac{dof} actuation by tilting both front and back rotors. While this design is effective, similar tasks can be performed by tilting only one side of the rotors (front or back), reducing the complexity of actuation, see Fig.~\ref{fig:design1}. This minimalist rotor configuration, however, raises challenges when significant forces need to be exerted on the environment.}

\update{Existing approaches to increase the force generation of aerial platforms typically involve increasing thrust capacity through larger propellers, co-axial designs \cite{bodie2019}, perching \cite{sun}, or adding mechanical compliance \cite{bart}. Additionally, most current designs feature a fixed \ac{com} within the rotor-defined area. In \cite{fabio}, a moving battery was used to compensate for the gravitational effects of a heavy manipulation arm. However, there remains a gap in the literature concerning the effect of various \ac{com} locations on force generation during physical interactions.} 

\update{In this work, we address this gap by proposing a novel aerial manipulation platform that leverages dynamic \ac{com} displacement to enhance force generation during physical interactions. In contrast to \cite{tmech}, we provide a comprehensive analysis of the platform’s force generation as influenced by varied \ac{com} locations. We detail the system’s design, modeling, and control and validate the proposed concepts through physical experiments involving both free-flight and pushing tasks. Additionally, we introduce a quantitative factor to compare the force exertion capabilities of our system with existing platforms, highlighting the advantages of our approach.}

\update{\section{Design Motivation}}\label{sec:motivation}
\update{In this section, we explore how varying the \ac{com} location affects force exertion in aerial manipulation, which motivates the design of our proposed aerial system. Specifically, we analyze the performance of an H-shaped aerial system equipped with tilting back rotors and fixed front rotors, as illustrated in Fig.~\ref{fig:design1}. This system features a rigid link attached to the vehicle body for interacting with a vertical work surface during a pushing task. Our analysis focuses on how the \ac{com}'s position influences the system’s ability to exert forces effectively on the work surface.}
\update{\subsection{Wrench Analysis}}\label{sec:analysis}
\begin{figure}[!t]
   \centering
\includegraphics[trim={0.8cm 0.5cm 0.7cm  0.5cm},clip,width=0.7\columnwidth]{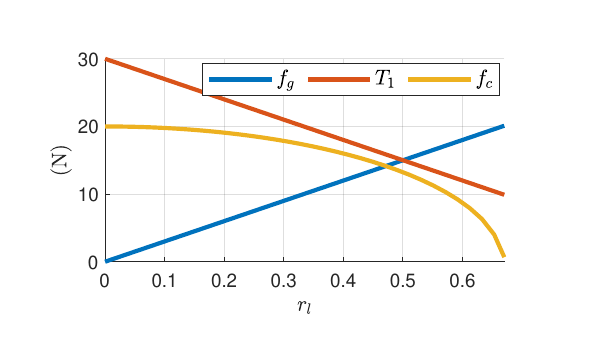}
    \caption{\update{The variation of forces $f_c$ and $f_g$ \ac{wrt} $r_l$ based on \cref{eq:force_com,eq:force_com2,eq:force_com3} is depicted.}}
    \label{fig:force_com}
\end{figure}
\update{We assume that the aerial vehicle is symmetric around the main axis of the rigid link along the interaction direction as in Fig.~\ref{fig:design1}. By neglecting small friction forces, the problem simplifies to a planar scenario, as illustrated in the free body diagram in Fig.~\ref{fig:design2}. We denote $T_1$ and $T_2$ as the thrust magnitude of the front and back rotors, respectively, and $G_0$ as the gravitational force acting on the total system. The component of $T_2$ along the interaction direction is $f_h$, while $f_g$ represents the component of $T_2$ opposing the gravitational force. The contact force exerted by the environment, passing through the system's \ac{com}, is denoted by $f_c$. We define $l_1$ and $l_2$ as the distances between the propellers' centers and the \ac{com}. Assuming the system is in equilibrium during interaction, with zero net forces and torques, the following relationships hold:
\begin{align}\label{eq:force_com}
    f_g&=\frac{G_0l_1}{l_1+l_2}=G_0r_l,\\
    T_1&=G_0(1-r_l)\label{eq:force_com2},\\
    f_c&=f_h=\sqrt{T_2^2-f_g^2}\label{eq:force_com3}.
\end{align}
where $r_l=\displaystyle\tfrac{l_1}{l_1+l_2} \in [0,1]$. The distance between front and back rotors $l_1+l_2$ is fixed. To investigate the effect of \ac{com} location on the system's force exertion, we analyze the case with $G_0=30\si{\newton}$, $T_2=20\si{\newton}$, and $l_1+l_2=0.27\si{\meter}$. By varying $r_l$ within its feasible range, the forces change according to \cref{eq:force_com,eq:force_com2,eq:force_com3}, as depicted in Fig.~\ref{fig:force_com}.}

\update{In traditional aerial system designs with a fixed \ac{com}, the configuration typically corresponds to $r_l=0.5$. In this setup, the thrust from the back rotor generates a component $f_g$ that partially compensates for the gravitational effect, in conjunction with the thrust from the front rotor $T_1$. Consequently, the force applied to the environment $f_h=f_c$ is limited to $\frac{13}{20}=0.65$ of the back rotor thrust $T_2$. When the \ac{com} is shifted closer to the back rotor ($r_l>0.5$), the force exerted on the environment decreases. As illustrated in Fig.~\ref{fig:force_com}, $f_c$ approaches zero as $r_l$ increases beyond $0.5$, reaching near zero around $r_l=0.7$. This demonstrates that moving the \ac{com} closer to the back rotor reduces the system’s overall force exertion capability.}

\update{Conversely, moving the \ac{com} closer to the front rotors ($r_l<0.5$) enhances force generation on the work surface. When $r_l=0$, the system exerts the maximum force on the environment, with $f_c=T_2$. In this configuration, the back rotor thrust is fully utilized to generate the contact force on the work surface, while the gravitational component $f_g$ is zero. The system's \ac{com} is aligned with the front rotors at $l_1=0$ and the front rotor thrust fully compensates for the gravitational force, resulting in $T_1=G_0$.}

\update{\subsection{ Proposed Design Concept}\label{sec:concept}}
\update{The wrench analysis indicates a novel approach to enhance force generation in aerial systems: shifting the system's \ac{com} closer to the front rotors. Traditional underactuated aerial vehicles with 4-\ac{dof} are known for their energy efficiency, particularly when carrying heavy loads or manipulators. Therefore, we propose a novel aerial system design that functions as a 4-\ac{dof} vehicle during navigation (in navigation mode), with a fixed \ac{com} aligned with the \ac{cog}.}

\update{For interaction tasks (in interaction mode), the platform features tiltable back rotors and a dynamic \ac{com} displacing configuration. This configuration allows the platform to adjust its \ac{com} position closer to the front rotors during interactions, thereby enabling higher force exertion.}

\update{In the following sections, we detail the design, modeling, and control of this aerial manipulation system, specifically addressing the dynamic \ac{com} design.}

%% file: texts/02_system_design.tex
\section{System Design}\label{sec:system_design}
\begin{figure}[!t]
   \centering
\includegraphics[width=0.95\columnwidth]{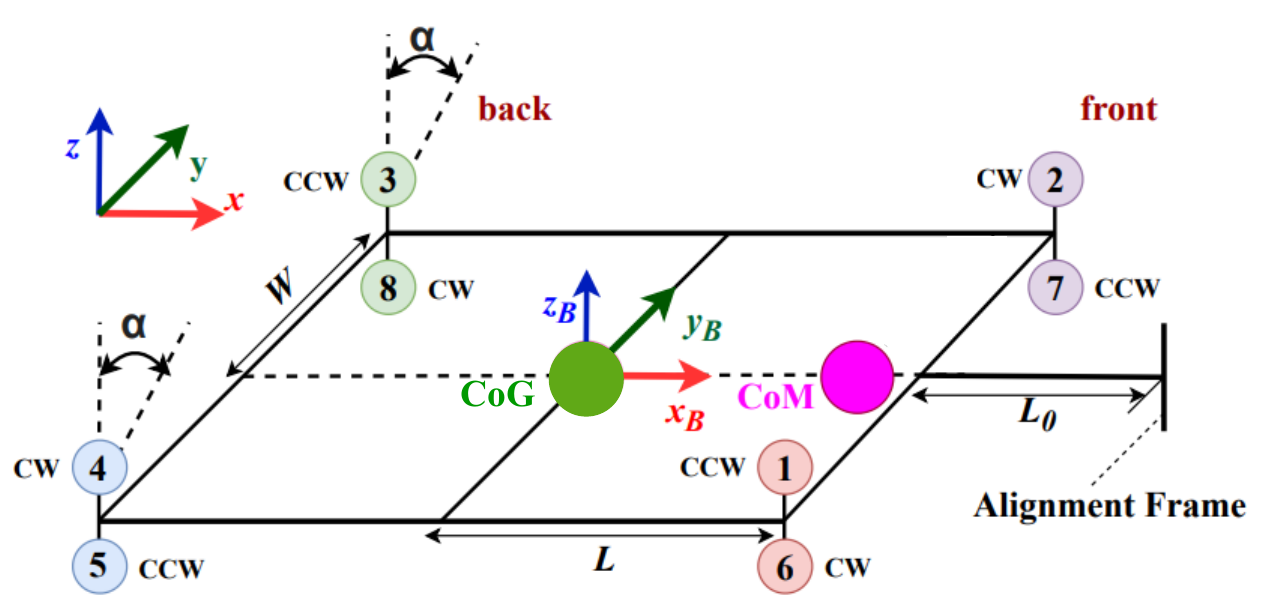}
    \caption{The world frame $\mathcal{F}_w=\{\bm{O};\bm{x},\bm{y},\bm{z}\}$ and the body frame $\mathcal{F}^B$. The rotors 3, 4, 5, and 8 can tilt simultaneously with $\alpha \in [-90\degree, \ 90\degree]$ around $\bm{y}_B$; the system's \ac{com} can shift along $\bm{x}_B$. CCW: counter-clockwise; CW: clockwise.}
    \label{fig:system}
\end{figure}
In this section, we detail the system design of the aerial vehicle \update{adapting the design concept introduced in the previous section.
The aerial vehicle's rotor configuration follows the design of an H-shaped coaxial octocopter with the body frame $\mathcal{F}^B=\{O_B;\bm{x}_B,\bm{y}_B,\bm{z}_B\}$ attached to its \ac{cog}, as in Fig.~\ref{fig:system}. Rotors $1$, $2$, $6$, and $7$ in the front have fixed rotating axes parallel to $\bm{z}_B$. Rotors $3$, $4$, $5$, and $8$ on the back have parallel tiltable axes that can simultaneously tilt around $\bm{y}_B$ via a servo motor. We denote $\alpha \in [-90\degree, \ 90\degree]$ as the tilting angle between each tiltable axis and $\bm{z}_B$. 
Such design introduces an additional \ac{dof} to the system compared to the classic coaxial octocopters with $4$-\ac{dof} actuation by entailing the decoupling between horizontal force generation along $\bm{x}_B$ and gravity force compensation of the platform. This feature enables interactions with non-horizontal surfaces at different orientations using a rigidly attached link, referred to as the \textit{alignment frame}, see Fig.~\ref{fig:physical_model}.}

\update{The aerial vehicle also features a dynamic \ac{com} that can be displaced along the body axis $\bm{x}_B$. This displacement is achieved by moving a shifting-mass plate, which includes heavy components such as the battery, manipulator, and tools, using a linear actuator. Positioning the shifting-mass plate at the aerial vehicle's \ac{cog} ensures that the system’s initial \ac{com} coincides with the \ac{cog}. Positioning the shifting-mass plate at the aerial vehicle's \ac{cog} allows us to assume that the system's initial \ac{com} coincides with the \ac{cog}. The distance between the system \ac{cog} and each rotor center along $\bm{x}_B$ is represented by $L$, while the distance along $\bm{y}_B$ is denoted as $W$. $L_0$ represents the length of the alignment frame.}

\begin{figure}[!t]
   \centering
\begin{subfigure}{0.52\columnwidth}\centering
    \includegraphics[trim={0.7cm 0.4cm 0.55cm  0.6cm},clip,width=\columnwidth]{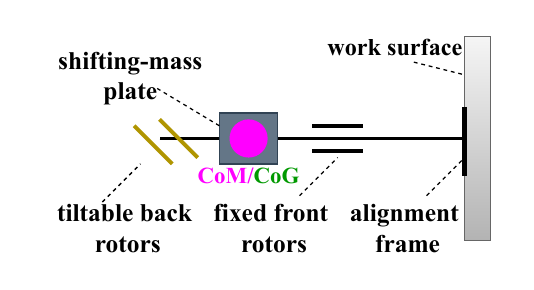}
    \caption{}
    \label{fig:action1}
\end{subfigure}
     \begin{subfigure}{0.46\columnwidth}\centering
    \includegraphics[trim={0.5cm 0.1cm 0.55cm  0.0cm},clip,width=\columnwidth]{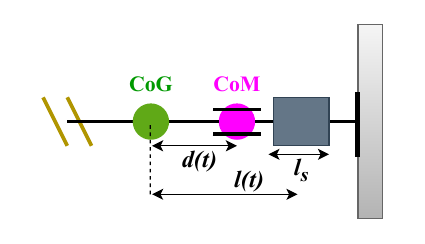}
    \caption{}
    \label{fig:action2}
\end{subfigure}
 \begin{subfigure}{\columnwidth}\centering
    \includegraphics[trim={0.5cm 0.1cm 0.55cm  0.0cm},clip,width=\columnwidth]{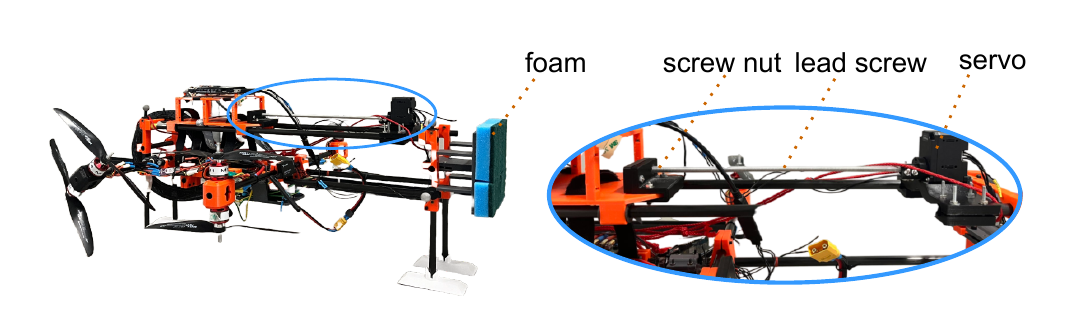}
    \caption{}
    \label{fig:detail_model}
\end{subfigure}
\caption{\update{(a-b) Schematics illustrating the designed aerial vehicle's operational flow for pushing on a vertical surface: (a) The alignment frame allows the platform to interact with the work surface.~(b) The shifting-mass plate changes its position $l(t)$ with the resulted dynamic system \ac{com} displacement $d(t)$ while the alignment frame is preserving the contact. (c) A physical prototype of the designed aerial vehicle and the zoomed-in imagine of the mechanism for moving the shifting-mass plate.}}\label{fig:aero_bull}
\end{figure}

\update{As outlined in Sec.~\ref{sec:concept}, different operation modes are assigned to the aerial vehicle including the navigation mode and the interaction mode for pushing tasks. In navigation mode, the vehicle functions as a classic coaxial octocopter with a fixed \ac{com}. Upon switching to interaction mode, the tiltable back rotors and the dynamic \ac{com} become active. The platform establishes initial contact with the work surface using the alignment frame, with the \ac{com} positioned as shown in Fig.~\ref{fig:action1}. While maintaining this contact, the shifting-mass plate moves closer to the work surface, thereby bringing the \ac{com} nearer to the front rotors, as illustrated in Fig.~\ref{fig:action2}. As the \ac{com} shifts closer to the front rotors (with a reduced $r_l$), more thrust from the back rotors can be utilized to generate a higher pushing force as depicted in Fig.~\ref{fig:force_com}.} 

\subsection{Dynamic Center-of-Mass Displacing}\label{sec:mass_shift}
We denote $m_S, \ m \in \mathbb{R}^+$ as the mass of the shifting-mass plate and the total system respectively, where $m_S<m$. The shifting-mass plate is connected to a linear actuator and can move along $\bm{x}_B$ starting from the system's \ac{cog} up to the alignment frame tip. 
The location of the shifting-mass plate is represented by its \ac{cog} position. We denote $l_S$ as the length of the shifting-mass plate along the body axis $\bm{x}_B$, as in Fig.~\ref{fig:action2}. The shifting-mass plate position along $\bm{x}_B$ \ac{wrt} the body frame origin $\bm{O}_B$ is denoted as $l(t) \in [0,L+L_0-0.5l_S]$. Changing the shifting-mass plate position results in a displacement of the system's \ac{com} along $\bm{x}_B$, denoted by $d(t)$. When the system's \ac{com} is located at $d>L$ outside the rotor-defined area (i.e., over-displaced), the aerial vehicle often flips around the contact area, leading to instability. Therefore, $d(t)$ is restricted by $[0, L]$ to avoid risky scenarios and damage to the platform. Assuming the symmetric mass distribution of the platform around its body axes, the relation between $l(t)$ and $d(t)$ is given by: 
\begin{equation}\label{eq:shift_com}
    d(t)=\frac{m_S}{m}l(t).
\end{equation}

\update{With \cref{eq:shift_com},the maximum \ac{com} displacement $d=L$ corresponding to the case $r_l=0$ in Fig.~\ref{fig:force_com} results in a shifting-mass plate position of $l=\displaystyle\tfrac{m}{m_S}L>L$. The shifting-mass plate position is thus restricted to $[0, \displaystyle\tfrac{m}{m_S}L]$.
Neglecting friction forces from the work surface, with $d=L$, only front rotors supply the force for gravity compensation. The back rotors can tilt up to $\alpha=90 \degree$ to be fully used for generating the pushing force.}
\update{\subsection{Physical Prototype}}
\update{A physical prototype with a total weight of $3.12\si{\kilogram}$ has been developed from scratch, utilizing 3D-printed parts and carbon-fiber tubes. This prototype facilitates simple pushing tasks using the alignment frame, as shown in Fig.~\ref{fig:detail_model}. The linear motion of the shifting-mass plate is enabled by a lead screw, a Dynamixel servo, and a screw nut housed in a box rigidly attached to the shifting-mass plate, for details please refer to \cite{tmech}. This design ensures precise, slow motion of the plate, with a velocity of $0.001\si{\meter\per\second}$, making the dynamics of shifting the \ac{com} negligible. To enhance the interaction quality, foam is attached to the tip of the alignment frame, providing mechanical compliance to reduce impact during initial contact and ensuring smooth interaction.}

%% file: texts/03_modeling.tex
\section{System Modeling}
\label{sec:modeling}
In this section, we present the system modeling of the designed aerial vehicle in Sec.~\ref{sec:system_design}. Concerning Fig.~\ref{fig:system}, we define  $\mathcal{F}_w=\{\bm{O};\bm{x},\bm{y},\bm{z}\}$ as the world frame. The thrust magnitude of the $i$-th rotor is denoted by $T_i=k_t\Omega_i^2 \in \mathbb{R}$, where $k_t \in \mathbb{R}$ is the thrust coefficient and $\Omega_i \in \mathbb{R}^+$ is the rotor rotating speed. According to the rotor configuration in Fig.~\ref{fig:system}, the drag torque of the $i$-th rotor along the positive thrust direction is given by $\tau_i =(-1)^i k_b \Omega_i^2 \in \mathbb{R}$, where $k_b \in \mathbb{R}$ is the drag torque coefficient. \update{The rotation matrix $\bm{R}(\phi, \theta, \psi) \in SO(3)$ represents the orientation of the body frame $\mathcal{F}_B$ w.r.t. the world frame
where $\phi$, $\theta$, $\psi \in (-90\degree,90\degree)$ are roll-pitch-yaw Euler angles.}
\subsection{Moment of Inertia Estimation} \label{sec:inertia}
To simplify the modeling, we assume that the body axes of frame $\mathcal{F}_B$ correspond with the system's principal axes of inertia. 
Unlike the conventional aerial vehicles with fixed centers of mass and constant moments of inertia, while displacing the system's \ac{com} along the body axis $\bm{x}_B$, the moments of inertia along body axes $\bm{y}_B$ and $\bm{z}_B$ vary.

To reduce modeling inaccuracies, we present the moment of inertia estimation associated with the shifting-mass plate position $l(t)$. 
To do so, the geometry, mass, and moments of inertia of the main components used in the physical model in Fig.~\ref{fig:physical_model} are captured by realistic CAD models. By measuring, $I_{xx}=0.0444$\si{\kilo\gram\square\meter} stays constant for different value of $l$, while $I_{yy}$ and $I_{zz}$ change values along with $l$. With the measured data and similarly to \cite{aero_tong}, linear regression technology is applied to obtain mathematical models of $I_{yy}$ and $I_{zz}$ \ac{wrt} $l$:
\begin{equation}\label{eq:inertia}
\begin{split}
    &I_{yy}(l)=0.49l^2+0.0538 \ (\si{\kilo\gram\square\meter}),\\
    &I_{zz}(l)=0.52l^2+0.0795 \ (\si{\kilo\gram\square\meter}).
    \end{split}
\end{equation}
The estimated inertia matrix of the designed system is thus given by $\bm{I}(l)=diag(\begin{bmatrix}
    I_{xx}&I_{yy}(l)&I_{zz}(l)
\end{bmatrix})\in\mathbb{R}^{3\times3}$.
\subsection{System Actuation}\label{sec:actuation}
We now introduce the actuation wrenches (i.e., forces and torques) of the system in Fig.~\ref{fig:system}. Given the tiltable back rotors, the aerial vehicle can have thrust components along the interaction axis $\bm{x}_B$. The actuation force vector $\bm{F}_a \in \mathbb{R}^{3}$ expressed in the body frame $\mathcal{F}_B$ is given by\footnote{For simplicity, $C_{(\cdot)}$ and $S_{(\cdot)}$ denote $\cos(\cdot)$ and $\sin(\cdot)$, while $(\cdot)_{1,2,3,...,n}$ represents the sum of thrust and drag torques, e.g., $T_{3,4,5,8}=T_3+T_4+T_5+T_8$}:
\begin{equation}\label{eq:f_a}
    \bm{F}_a=\begin{bmatrix}
        F_1\\0\\F_3
    \end{bmatrix}=\begin{bmatrix}
        T_{3,4,5,8}S_{\alpha}\\0\\T_{3,4,5,8}C_{\alpha}+T_{1,2,6,7}
    \end{bmatrix}.
\end{equation}
The actuation torque vector $\bm{\Gamma}_a \in \mathbb{R}^{3}$ \ac{wrt} the system's \ac{com} expressed in $\mathcal{F}_B$ is given by:
\update{
\begin{equation}\label{eq:t_a}
    \resizebox{0.9\columnwidth}{!}{ $\bm{\Gamma}_a=\begin{bmatrix}
        \Gamma_1\\\Gamma_2\\\Gamma_3
    \end{bmatrix}=\begin{bmatrix}
        \big(T_{2,7}-T_{1,6}+(T_{3,8}-T_{4,5})C_{\alpha}\big)W+\tau_{3,4,5,8}S_{\alpha}\\T_{3,4,5,8} C_{\alpha}L-T_{1,2,6,7}L\\ (T_{4,5}-T_{3,8})S_{\alpha}W+\tau_{3,4,5,8}C_{\alpha}+\tau_{1,2,6,7}
    \end{bmatrix}$}.
\end{equation}}
The system therefore has 5-\ac{dof} actuation provided by 9 system inputs including the rotors' rotating speed and back rotors' tilting angle. We denote $\bm{u}=\begin{bmatrix}
    \alpha&\Omega_1&...&\Omega_8
\end{bmatrix}^{\top} \in \mathbb{R}^9$ as the system inputs.
\subsection{Equations of Motion}
Assuming slow motion of the shifting-mass plate, we neglect the dynamics of the moving plate. The system equations of motion expressed in the body frame $\mathcal{F}_B$ can be written as:
\update{\begin{equation}
         \begin{bmatrix} \bm{M} & \bm{0} \\ \bm{0} &\bm{I}\end{bmatrix} \begin{bmatrix}
             \dot{\bm{\upsilon}} \\\dot{\bm{\omega}}  
         \end{bmatrix}+\begin{bmatrix}
             \mathbb{S}(\bm{\omega}) \bm{M}\bm{\upsilon} \\  \mathbb{S}(\bm{\omega}) \bm{I}\bm{\omega}  \end{bmatrix}+\bm{G}=\begin{bmatrix}
                 \bm{F}_a \\ \bm{\Gamma}_a
             \end{bmatrix}+ \begin{bmatrix}
                 \bm{F}_C \\ \bm{\Gamma}_C
             \end{bmatrix}.
             \label{eq:dyn_model}
\end{equation}}
where $\bm{M} \in \mathbb{R}^{3\times 3}$ is the mass matrix. \update{$\bm{G} \in \mathbb{R}^{6}$ is the gravity term expressed in the body frame. Considering the \ac{com} displacing, we have:
\begin{equation}
    \resizebox{0.8\columnwidth}{!}{$\bm{G}(l)=\begin{bmatrix} \bm{R}^{\top}g\bm{e}_3\\\mathbb{S} (\bm{R}^{\top}g\bm{e}_3)\begin{bmatrix}
        -d(t)&0&0
    \end{bmatrix}^{\top}\end{bmatrix}=\begin{bmatrix}
        \bm{G}_{lin} \\ \bm{G}_{ang}
    \end{bmatrix}$},
\end{equation}
with $g=-9.81$\si{\meter\per\square\second}, $\bm{e}_3=\begin{bmatrix}
    0&0&1
\end{bmatrix}^{\top}$, and $d=\frac{m_S}{m}l$ from \cref{eq:shift_com}.} $\bm{\upsilon},\bm{\omega} \in \mathbb{R}^{3}$ are the linear and angular velocity vectors of the body frame $\mathcal{F}_B$ \ac{wrt} the world frame. Assuming that the interaction wrenches at the end-effector and alignment frame tip are the only sources of external wrenches, we denote $\begin{bmatrix} \bm{F}_C \\ \bm{\Gamma_C} \end{bmatrix} \in \mathbb{R}^{6}$ as the stacked force and torque vector exerted from the environment to the system. We use $\mathbb{S}(\bm{a}) \in \mathbb{R}^{3\times3}$ to present the skew symmetric matrix such that $\mathbb{S}(\bm{a})\bm{b}=\bm{a}\times \bm{b}$, and we have $\dot{\bm{R}}=\bm{\omega}\times\bm{R}=\mathbb{S}(\bm{\omega})\bm{R}$. We define the operator $(\cdot)^{\vee}$ such that $\bm{a}=\mathbb{S}(\bm{a})^{\vee}$.

%% file: texts/04_control.tex
\begin{figure}[!t]
   \centering
\includegraphics[width=0.9\columnwidth]{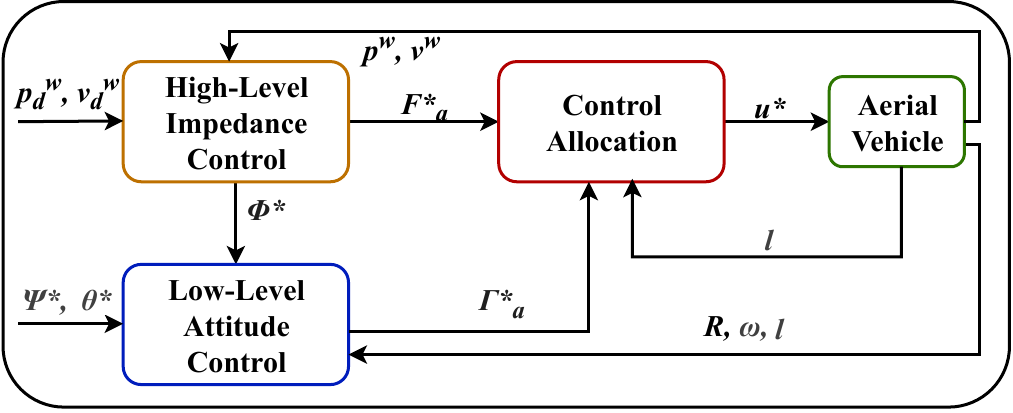}
    \caption{Overall control scheme: the main blocks are depicted.}
    \label{fig:control}
\end{figure}  
\section{Control Design}\label{sec:control}
\update{In this section, we present the control design of the aerial vehicle for the interaction mode with a dynamic \ac{com} displacing configuration. Considering the modeled system's under-actuation and inspired by \cite{ding, bodie2019, lee2013}, we employ a cascade control pipeline as summarized in Fig.~\ref{fig:control} 
with the capability to handle both free flights and physical interactions. In the following, $(\cdot)^*$ 
 indicate the desired quantities. 
This control framework integrates low-level geometric attitude control and high-level selective impedance control outputting the desired actuation wrenches $F_a^*$, $\Gamma_a^*\in\mathbb{R}^3$. These outputs are sent to the control allocation to compute the desired rotors' rotating speed $\Omega_i^*$ and tilting angle $\alpha^*$.}
\subsection{Low-level Geometric Attitude Control}\label{sec:attitude_control}
\update{We use the geometric controller in~\cite{lee2013} to control the system's attitude dynamics. The low-level controller outputs the desired actuation torques $\bm{\Gamma}_a^*$ by tracking the system's orientation and angular velocity in the body frame. The attitude tracking error $\bm{e}_R \in \mathbb{R}^3$ is defined as
$
    \bm{e}_R=\frac{1}{2}(\bm{R}^{*\top}\bm{R}-\bm{R}^{\top}\bm{R}^*)^{\vee},
$
and the angular velocity tracking error $\bm{e}_{\omega} \in \mathbb{R}^3$ is defined as:
$
    \bm{e}_{\omega}=\bm{\omega}-\bm{R}^{\top}\bm{R}^* \bm{\omega}^*
$. We define the integral error $\bm{e}_I=\int_o^t \bm{e}_{\omega}(\tau)+c_2\bm{e}_R(\tau)d\tau$, with $c_2$ being a positive constant.
A nonlinear attitude dynamics feedback controller is designed as the following:
\begin{equation}
    \begin{split}
    \bm{\Gamma}_a^*(l)=&-\bm{K}_R \bm{e}_R-\bm{K}_{\omega}\bm{e}_{\omega}-\bm{K}_I\bm{e}_I+\bm{\omega} \times \bm{I}(l)\bm{\omega}\\
    &+\bm{G}_{ang}-\bm{I}(l)(\mathbb{S}(\bm{\omega})\bm{R}^{\top}\bm{R}^*\bm{\omega}^*-\bm{R}^{\top}\bm{R}^*\dot{\bm{\omega}}^*),
\end{split}\label{eq:torque}
\end{equation}
where $\bm{K}_R$, $\bm{K}_{\omega}$, and $\bm{K}_I \in \mathbb{R}^{3\times3}$ are positive-definite matrices.} 
\subsection{High-level Selective Impedance Control}\label{sec:impedance}
\update{The high-level selective impedance control outputs the desired actuation forces $\bm{F}_a^*$ by tracking the body frame origin $O_B$ attached to the aerial vehicle's \ac{cog}, instead of the shifting-mass plate in the world frame. The system's linear dynamics can be re-written as:
\begin{equation}\label{eq:euqation_moton_w}
    \bm{M}\dot{\bm{\upsilon}}^w+\mathbb{S}(\bm{\omega}^w) \bm{M}\bm{\upsilon}^w+mg\bm{e}_3=\bm{F}_a^w+\bm{F}_C^w,
\end{equation}
where $\bm{F}_a^w=\bm{R}\bm{F}_a$ and $\bm{F}_C^w=\bm{R}\bm{F}_C$. 
Considering the frame transformation and  \cref{eq:f_a}, $\bm{F}_a^w$ is given by:
\begin{equation}\label{eq:f_a_w}
   \bm{F}_a^w=\begin{bmatrix}
        C_{\theta}C_{\psi}F_1+ (S_{\theta}C_{\psi}C_{\phi}+S_{\psi}S_{\phi})F_3 \\ C_{\theta}S_{\psi}F_1+(S_{\theta}S_{\psi}C_{\phi}-C_{\psi}S_{\phi})F_3\\
        -S_{\theta}F_1+ C_{\theta}C_{\phi}F_3
    \end{bmatrix}.
\end{equation}}

\update{The position tracking error is given by  
$
\bm{e}_p^w=\bm{p}^w-\bm{p}^{w*}
$, 
where $\bm{p}^w\in \mathbb{R}^3$ represents the position of the origin $O_B$ \ac{wrt} $O$ expressed in the world frame. The velocity tracking error can then be displayed as $\bm{e}_v^w=\bm{\upsilon}^w-\bm{\upsilon}^{w*}$ with $\bm{\upsilon}^{w*}=\dot{\bm{p}}^{w*}$. The desired close loop dynamics of the system with the selective impedance control is:
\begin{equation}\label{eq:close_loop}
    \bm{M}\dot{\bm{e}}_v^w+\bm{D}\bm{e}_v^w+\bm{K}\bm{e}_p^w=\bm{F}_C^w,
\end{equation}
where $\bm{D}$ and $\bm{K} \in \mathbb{R}^{3\times3}$ are positive-definite matrices representing the desired damping and stiffness of the system. 
Combining~\cref{eq:euqation_moton_w} and~\cref{eq:close_loop}, the desired actuation force vector expressed in the world frame is defined as:
\begin{equation}\label{eq:impedance}
\begin{split}
    \bm{F}_a^{w*}&=\begin{bmatrix}
       F_1^{w*}&F_2^{w*}&F_3^{w*}
\end{bmatrix}^{\top}\\
&=\bm{M}\dot{\bm{\upsilon}}^{w*}-\bm{D}\bm{e}_v^w-\bm{K}\bm{e}_p^w+\bm{\omega}^w\times \bm{M}\bm{\upsilon}^w+mg\bm{e}_3.
   \end{split}
\end{equation}}

\update{The system introduces coupling between its roll angle $\phi$ and its linear motion in the world frame not having a thrust component along the body axis $\bm{y}_B$. Again, combining \cref{eq:f_a_w} and \cref{eq:impedance} the desired roll angle $\phi^*$ is calculated via:
\begin{equation}\label{eq:desired_roll}
\begin{split}
    \phi^*=&atan2\big(S_{\psi}F_1^{w*}-C_{\psi}F_2^{w*},S_\theta C_\psi F_1^{w*}+...\\
    &S_\theta S_\psi F_2^{w*}+C_{\theta}F_3^{w*}\big).
    \end{split}
\end{equation}
 Furthermore, by using trigonometric calculations with~\cref{eq:f_a_w}, the desired actuation force vector $\bm{F}_a^*=\begin{bmatrix}
     F_1^*&0&F_3^*
 \end{bmatrix}^{\top}$ expressed in the body frame is given by:
 \begin{subequations}
 \begin{equation}
     F_1^*=C_\theta C_\psi F_1^{w*}+C_\theta S_\psi F_2^{w*}-S_\theta F_3^{w*},
     \end{equation}
     \begin{equation}
          \begin{split}
      F_3^*=&\big((S_{\psi}F_1^{w*}-C_{\psi}F_2^{w*})^2+(S_\theta C_\psi F_1^{w*}+...\\
      &S_\theta S_\psi F_2^{w*}+C_\theta F_3^{w*})^2\big)^{\frac{1}{2}}.
 \end{split}
     \end{equation}
 \end{subequations}
 The desired roll angle $\phi^*$ is then fed to the low-level attitude controller in Sec.~\ref{sec:attitude_control} with the attitude references $\psi^*$ and $\theta^*$. 
 Finally, $\bm{F}_a^*$ and $\bm{\Gamma}_a^*$ are fed to the control allocation to derive the desired system inputs $\bm{u}^*$, as in Fig.~\ref{fig:control}.}
 \subsection{Control Allocation}
\update{The control allocation problem for the examined case can be divided into finding the desired tilting angle $\alpha^*$ value,
\begin{equation}\label{eq:alpha}
    \alpha^*=\frac{\pi}{2}-atan2\big(F_3^*L+\Gamma_2^*(l),2LF_1^*\big),
\end{equation}
and retrieve the corresponding motor velocities $\Omega_i$. To handle the redundancy of the system, we introduce a virtual control input vector $\bm{\lambda}=\begin{bmatrix}
    \Omega_1^2 & \Omega_2^2 & ... & \Omega_8^2\end{bmatrix}^\top \in \mathbb{R}^{8}$. By applying~\cref{eq:shift_com},~\cref{eq:f_a}, and~\cref{eq:t_a} the desired virtual control vector is calculated via pseudo-inverse:
\begin{equation}\label{eq:lambda}
    \bm{\lambda}^*=\bm{H}(\alpha^*)^{\dag}\cdot  \begin{bmatrix}
    \bm{F}_a^*\\\bm{\Gamma}_a^*
    \end{bmatrix},
\end{equation}
where $\bm{H}(\alpha^*) \in \mathbb{R}^{6\times12}$ is the allocation matrix depending on the desired tilting angle $\alpha^*$ and $(\cdot)^{\dag}$ is the associate pseudo-inverse operator. Finally, we can compute the desired real system inputs as $\bm{u}^*=\begin{bmatrix} \alpha^* & \sqrt{\bm{\lambda}^*} \end{bmatrix}$. This approach minimizes the motors' rotating speed with guaranteed positive values~\cite{mina}.}

%% file: texts/06_experiment.tex
\section{experimental results}\label{sec:real_test}
\begin{figure*}[!t]
    \centering
    \subfloat[]{\label{fig:real_tests_lin}\includegraphics[width = 0.48\linewidth]{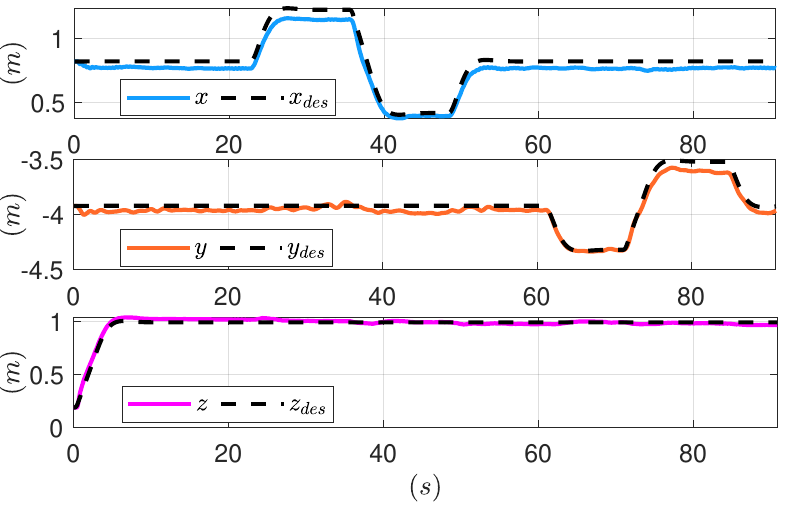}}
    \subfloat[]{\label{fig:real_tests_att}\includegraphics[width = 0.49\linewidth]{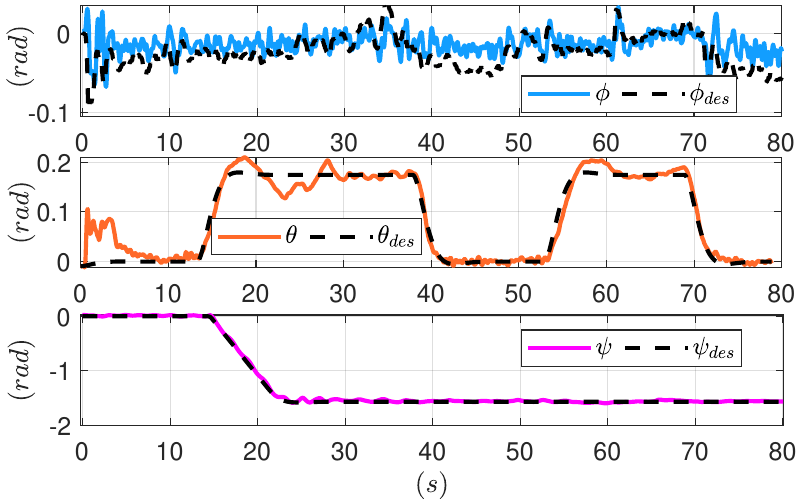}}
    \caption{(a) Position tracking. (b) Attitude tracking. Dashed lines represent the set points while continuous ones are the aerial vehicle pose feedback.}
\end{figure*}
\update{In this section, we present the experimental results obtained from the built system, including both free-flight tests and two interaction task scenarios. The platform’s navigation mode is managed by the standard PX4 firmware \cite{px4}, while the interaction mode is tested with a customized controller, as shown in Fig.~\ref{fig:control}. Free-flight tests were conducted to validate the performance of the proposed cascade controller described in Sec.~\ref{sec:control}. In \textbf{Scenario 1} and \textbf{Scenario 2}, we performed pushing tasks on a vertical surface with the platform both with and without dynamically displacing the \ac{com}. These scenarios are designed to demonstrate the advantages of the proposed system's dynamic \ac{com} configuration. Additionally, we introduce a quantitative factor to compare the force exertion capabilities of our system with those of existing platforms.
}

\subsection{Experiment Setup}
The entire control architecture is implemented via a customized PX4 firmware \cite{simone} enabling the support for the servo motor to tilt back rotors and a ROS2 controller (in C++) implementing the designed control law. A separate ROS2 controller is used to move the shifting-mass plate, in which the shifting-mass plate position $l$ is mapped into the servo position. The C++ control node operates at $250$~Hz matching the standard PX4 controller frequency. The following gain values are used during the experiments: $\bm{K}=diag([22.0, \ 22.0, \ 80.0])$, $\bm{D}=diag([10.0, \ 10.0, \ 45.0])$, $c_2=0.8$, $\bm{K}_R=diag([5.0, \ 5.0, \ 3.0])$, $\bm{K}_{\omega}=diag([1.0, \ 1.4, \ 0.25])$, and $\bm{K}_I=diag([0.0, \ 3.25, \ 0.5])$.
The codes run on an onboard Lattepanda Delta V3 which communicates with the flight controller via DDS and ROS2. An OptiTrack system is used to track the platform's position and orientation. A vertical wooden board is set as the work surface, and safety ropes are used to prevent platform damage, see Fig.~\ref{fig:physical_model}. 
\update{In practice, the mass distribution across the platform is not symmetric around the body axes. Directly moving the shifting-mass plate to $l=\displaystyle\frac{m}{m_S}L$ as introduced in Sec.~\ref{sec:mass_shift} may lead to over-displaced \ac{com} resulting in risky situations for the platform. Therefore, to avoid platform damage due to over-displaced \ac{com}, we tested a maximum shifting-mass plate position of $0.18\si{\meter}\ll\frac{m}{m_S}L=0.48\si{\meter}$.}
\update{\subsection{\textbf{Free-Flight Tests}}}
The first experiments evaluated trajectory tracking capabilities in Cartesian space. Firstly, desired position set-points $x_{des}$, $y_{des}$, and $z_{des}$ were sent to the aerial vehicle, and the tracking errors are displayed in Fig.~\ref{fig:real_tests_lin}. Due to the under-actuation of the system, the desired roll set points $\phi_{des}$ were generated from the corresponding linear motion of the platform as in \cref{eq:desired_roll}. Later, desired pitch and yaw set points $\theta_{des}$ and $\psi_{des}$ were sent to the aerial vehicle while hovering. The attitude tracking errors are displayed in Fig.~\ref{fig:real_tests_att}. 
The testing results show the designed system's promising orientation and position control during free flight with the proposed control law.
\update{\subsection{\textbf{Scenario 1}}}
\update{In this scenario, we conducted a series of pushing tasks on a vertical work surface to evaluate the feasibility and effectiveness of the system's dynamic \ac{com} configuration. The high-level impedance controller in Sec.~\ref{sec:control} treats the system as a mechanical admittance. To perform the pushing task, the desired contact position (i.e., the setpoint) was intentionally set \enquote{inside} the work surface, as per the approach outlined in \cite{review}. We denote $\delta p$ as the distance from the work surface to the setpoint along the interaction axis $\bm{x}_B$, with $\delta p>0$ indicating positions \enquote{inside} the surface.} 

\begin{figure}[!t]
    \centering
    \includegraphics[trim={0.5cm 0.5cm 0.5cm 0.2cm},clip,width=\columnwidth]{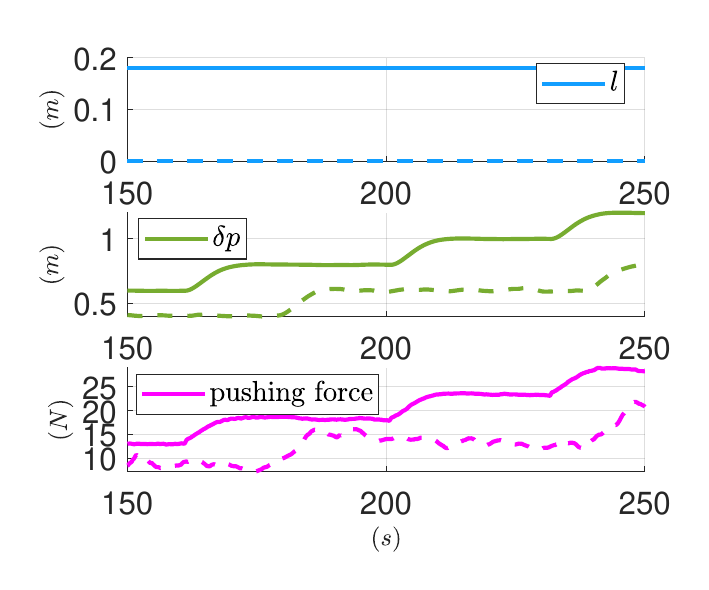}
    \caption{\update{From top to bottom: the shifting-mass plate position $l$, the setpoint position \ac{wrt} the work surface, and the pushing force normal to the work surface. Solid line: \textbf{Scenario 1} with displaced \ac{com} closer to front rotors. Dashed line: \textbf{Scenario 2} with non-displaced \ac{com}.}}
    \label{fig:reaction}
\end{figure}

\update{Using the experimental data, the pushing force perpendicular to the surface was estimated offline via the wrench estimation method detailed in \cite{Tomic2017,tong_aim}, utilizing the estimated thrust of each propeller. The required aerodynamic coefficients for thrust estimation were measured with an RC Benchmark Thrust Stand, with more details available in \cite{tmech}. To initiate contact with the surface, a setpoint of $\delta p=0.6\si{\meter}$ was provided to the high-level impedance controller, resulting in the platform exerting approximately $13\si{\newton}$ of pushing force, see Fig.~\ref{fig:reaction}. While maintaining contact, the shifting-mass plate moved actively toward the work surface, reaching its maximum allowable position at $l^*=0.18\si{\meter}$, thus dynamically displacing the \ac{com}. This process is demonstrated in the first part of the accompanying video\footnote[1]{\update{The video is available at \url{https://youtu.be/NXEnCS1ZLEg}.}.}} 

\update{With the shifting-mass plate at $l^*=0.18\si{\meter}$, setpoints of $\delta p=0.8, \ 1.0,\ 1.2\si{\meter}$ were then provided to command progressively higher pushing forces. The platform stably pushed against the vertical surface with forces of $19\si{\newton}$, $23\si{\newton}$, and $28\si{\newton}$, respectively (see Fig.~\ref{fig:reaction}). By moving the \ac{com} closer to the front rotors, as described in Sec.~\ref{sec:analysis}, the system was able to achieve stable pushing forces nearly equal to its gravitational force by tilting only the back rotors.}
\update{\subsection{\textbf{Scenario 2}}}
\update{Similar to \textbf{Scenario 1}, a series of setpoints with varying $\delta p$ values were used to conduct pushing tasks. However, in this case, the system's \ac{com} remained static, with the shifting-mass plate fixed at $l=0$, as depicted in Fig.~\ref{fig:reaction}. Throughout the tests, the platform exhibited more pronounced oscillations compared to \textbf{Scenario 1}, as shown in Fig.~\ref{fig:reaction} and the second part of the attached video\footnotemark[1]. Initially, a setpoint of $\delta p=0.2\si{\meter}$ was provided, resulting in an exerted pushing force of approximately $10\si{\newton}$. Subsequently, the platform applied a pushing force of about $15\si{\newton}$ with a setpoint of $\delta p=0.6\si{\meter}$. However, when the setpoint was increased to $\delta p=0.8\si{\meter}$, the system impulsively generated a force of $20\si{\newton}$, leading to risky conditions.}
\update{\subsection{Comparison with the existing platforms}}
\begin{table}[t!]
\centering
\caption{\update{Comparison with the SOTA.}}
\begin{tabularx}{\columnwidth}{p{0.09\textwidth}|p{0.08\textwidth}|p{0.04\textwidth}|p{0.12\textwidth}|X}
\toprule
\textbf{System} & \textbf{Mass (kg)} & \textbf{DoF} & \textbf{Num. of Tilting Axes for Pushing}&$h_f$ \\ \hline
s1(scenario 1)& $3.12$ & 5 &1 &0.92 \\ 
s1(scenario 2)& $3.12$ & 5 &1 &0.49 \\
s2 \cite{hwang} & $3.30$ & 5 & 2 &0.62 \\ 
s3 \cite{tong_voliro} & $4.00$ & 6 & 1 & 0.76\\ 
s4 \cite{bart} & 1.50& 4& 2&0.51\\
\hline
\end{tabularx}
\label{table_comp}
\end{table}
\update{In this section, we compare the force generation capability of the proposed system with various state-of-the-art (SOTA) aerial manipulation platforms. To evaluate force exertion on vertical surfaces during pushing tasks, we consider key parameters for each system: total system mass ($m_0$), the number of actuation \ac{dof}, the number of tilting rotor axes engaged for pushing ($n_t$), and the maximum achievable pushing force ($f_p$). Additionally, we introduce a comparison factor that accounts for both the platform's mass and the number of tilting axes used for pushing:
\begin{equation}
    h_f=\frac{f_p}{9.8\times m_0 n_t},
\end{equation}
where a higher value of $h_f$ indicates superior force exertion ability. Table \ref{table_comp} shows the comparison of our proposed system with SOTA platforms. System s2 \cite{hwang}, a 5-\ac{dof} platform, tilts both front and back rotors to generate pushing forces. It exerts a pushing force of $40\si{\newton}$ with a mass of $3.3\si{\kilogram}$ and two tilting axes. System s3 \cite{tong_voliro}, a 6-\ac{dof} platform, generates up to $30\si{\newton}$ of stable pushing force, with a mass of $4.0\si{\kilogram}$ and one tilting axis for pushing. System s4 \cite{bart}, featuring 4-\ac{dof} actuation, can apply up to $15\si{\newton}$ on a vertical surface with a weight of $1.5\si{\kilogram}$. It uses both front and back rotors for pushing, similar to the configuration in system s2 \cite{hwang}, with $n_t=2$.}

\update{In comparison, our proposed system, with only one tilting axis (the back rotors) and a lightweight structure ($3.12\si{\kilogram}$), can exert a stable pushing force of $28\si{\newton}$ when the \ac{com} is dynamically displaced closer to the front rotors as in \textbf{Scenario 1}. With a fixed \ac{com} as in \textbf{Scenario 2}, it can push up to $15\si{\newton}$ with more oscillations.}

\update{This quantitative comparison highlights the effectiveness of our approach, demonstrating that the proposed system achieves competitive force generation with fewer tilting axes thanks to the dynamic \ac{com} displacing configuration.}

%% file: texts/07_conclusion.tex
\section{Conclusion}\label{sec:conclusion}
\update{In this work, we introduced a novel approach to enhance force generation in aerial manipulation systems by focusing on the system’s \ac{com} location. Through an analysis of how the \ac{com} affects interaction forces, we proposed an innovative platform design featuring a dynamic \ac{com} displacing configuration. By shifting the \ac{com} closer to the front rotors during interactions, the system is capable of generating significantly higher pushing forces.
The effectiveness of this design was validated through physical experiments using an aerial vehicle equipped with tilting back rotors, achieving a stable pushing force nearly equal to the system's own weight. Additionally, we introduced a quantitative factor for comparing the force exertion capabilities of our system against existing platforms, demonstrating the clear advantages of our approach.
This work paves the way for the development of a new class of aerial vehicles with enhanced tool manipulation capabilities, opening up promising directions for future research and applications.}